\documentclass[11pt,a4paper]{article}
\usepackage[hyperref]{emnlp-ijcnlp-2019}
\usepackage{times}
\usepackage{latexsym}

\usepackage{url}

\usepackage{amsmath}
\usepackage{float}
\usepackage{graphicx}
\usepackage{longtable}
\usepackage[normalem]{ulem}

\usepackage{amsmath,amssymb,amsthm}
\usepackage{subcaption,siunitx,booktabs}
\usepackage{makecell}
\usepackage{arydshln}
\usepackage{xspace}
\usepackage{soul}
\usepackage{color}
\usepackage{multirow, booktabs}
\usepackage{bbding}
\usepackage{pifont}

\newcommand{\newvec}[1]{\mathbf{#1}}
\DeclareMathOperator*{\softmax}{softmax}

\newcommand{\sparc}{SParC\ }
\newcommand{\syncon}{SyntaxSQL-con}

\newcommand{\seqcon}{CD-Seq2Seq}
\newcommand{\xmark}{\ding{55}}

\newcommand{\keyword}[1]{\textcolor{blue}{#1}}
\newcommand{\sql}[1]{\textbf{\texttt{#1}}}

\newcommand{\hide}[1]{}

\aclfinalcopy 

\title{Editing-Based SQL Query Generation for \\ Cross-Domain Context-Dependent Questions}

\author{
Rui Zhang$^\dagger$
\quad Tao Yu$^\dagger$
\quad He Yang Er$^\dagger$
\\{\bf
\quad Sungrok Shim$^\dagger$
\quad Eric Xue$^\dagger$
\quad Xi Victoria Lin$^\P$
\quad Tianze Shi$^\S$}
\\{\bf
\quad Caiming Xiong$^\P$
\quad Richard Socher$^\P$
\quad Dragomir Radev$^\dagger$}\\
$^\dagger$ Yale University
\quad $^\P$ Salesforce Research
\quad $^\S$ Cornell University\\
\tt{\{r.zhang, tao.yu, dragomir.radev\}@yale.edu}\\
\tt{\{xilin, cxiong, rsocher\}@salesforce.com}
}

\date{}

\begin{document}
\maketitle
\begin{abstract}
We focus on the cross-domain context-dependent text-to-SQL generation task.
Based on the observation that adjacent natural language questions are often linguistically dependent and their corresponding SQL queries tend to overlap, we utilize the interaction history by editing the previous predicted query to improve the generation quality.
Our editing mechanism views SQL as sequences and reuses generation results at the token level in a simple manner.
It is flexible to change individual tokens and robust to error propagation.
Furthermore, to deal with complex table structures in different domains, we employ an utterance-table encoder and a table-aware decoder to incorporate the context of the user utterance and the table schema.
We evaluate our approach on the SParC dataset and demonstrate the benefit of editing compared with the state-of-the-art baselines which generate SQL from scratch.
Our code is available at \url{https://github.com/ryanzhumich/sparc_atis_pytorch}.
\end{abstract}

\section{Introduction}
Generating SQL queries from user utterances is an important task to help end users acquire information from databases.
In a real-world application, users often access information in a multi-turn interaction with the system by asking a sequence of related questions.
As the interaction proceeds, 
the user often makes reference to the relevant mentions 
in the history or omits previously conveyed information assuming it is known to the system.

Therefore, in the context-dependent scenario, the contextual history is crucial to understand the follow-up questions from users, and the system often needs to reproduce partial sequences generated in previous turns.
Recently, \newcite{suhr2018learning} proposes a context-dependent text-to-SQL model including an interaction-level encoder and an attention mechanism over previous utterances.
To reuse what has been generated, they propose to copy complete segments from the previous query.
While their model is successful to reason about explicit and implicit references, it does not need to explicitly address different database schemas because the ATIS contains only the flight-booking domain.
Furthermore, the model is confined to copy whole segments which are extracted by a rule-based procedure, limiting its capacity to utilize the previous query when only one or a few tokens are changed in the segment.

\begin{table*}[ht!]
\centering
\resizebox{\textwidth}{!}{
\begin{tabular}{cccccccccc}
\Xhline{4\arrayrulewidth}
      & Context & Cross-Domain & Interaction~(train~/~dev~/~test) & Question & Turn & Database & Table & Q. Length & Q. Vocab \\ \hline
Spider & \xmark & \checkmark & 11,840~(8,659~/~1,034~/~2,147) & 11,840 & 1.0 & 200 & 1020 & 13.4 & 4,818 \\
ATIS   & \checkmark & \xmark & 1,658~(1,148~/~380~/~130)           & 11,653      & 7.0          & 1           & 27       & 10.2 & 1,582  \\
SParC & \checkmark & \checkmark & 4,298~(3,034~/~422~/~842)           & 12,726      & 3.0          & 200         & 1020     & 8.1             & 3,794      \\
\Xhline{4\arrayrulewidth}
\end{tabular}
}
\caption{Dataset Statistics.}
\vspace{-2mm}
\label{tab:dataset}
\end{table*}

\begin{table*}[ht!]
\centering
\resizebox{.7\textwidth}{!}{
\begin{tabular}{ccccccccc}
\Xhline{4\arrayrulewidth}
      & \texttt{WHERE} & \texttt{AGG} & \texttt{GROUP} & \texttt{ORDER} & \texttt{HAVING} & \texttt{SET} & \texttt{JOIN} & \texttt{Nested} \\ \hline
Spider & 55.2 & 51.7 & 24.0 & 21.5 & 6.7 & 5.8 & 42.9 & 15.7 \\
ATIS   & 100      & 16.6   & 0.3      & 0        & 0         & 0      & 96.6    & 96.6      \\
SParC  & 42.8     & 39.8   & 20.1     & 17.0     & 4.7       & 3.5    & 35.5    & 5.7       \\
\Xhline{4\arrayrulewidth}
\end{tabular}
}
\caption{\% of SQL queries that contain a particular SQL component.}
\vspace{-3mm}
\label{tab:dataset_2}
\end{table*}

\begin{table*}[t!]
\begin{tabular}{|p{\textwidth}|}
\hline
\small{Database: student dormitory containing 5 tables.}\\
\small{Goal: Find the first and last names of the students who are living in the dorms that have a TV Lounge as an amenity.}\\
\hdashline
\small{Q1: How many dorms have a TV Lounge?}\\
\small{S1:} \scriptsize{\sql{\keyword{SELECT} \keyword{COUNT}(*) \keyword{FROM} dorm \keyword{AS} T1 \keyword{JOIN} has\_amenity \keyword{AS} T2 \keyword{ON} T1.dormid = T2.dormid \keyword{JOIN} dorm\_amenity \keyword{AS} T3}}\\
\vspace{-.18in}
~~~~~\scriptsize{\sql{\keyword{ON} T2.amenid = T3.amenid \keyword{WHERE} T3.amenity\_name = `TV Lounge'}}
\\
\small{Q2: What is the total capacity of these dorms?}\\
\small{S2:} \scriptsize{\sql{\keyword{SELECT} \keyword{SUM}(T1.student\_capacity) \keyword{FROM} dorm \keyword{AS} T1 \keyword{JOIN} has\_amenity \keyword{AS} T2 \keyword{ON} T1.dormid = T2.dormid \keyword{JOIN}}} \\
\vspace{-.18in}
~~~~~\scriptsize{\sql{dorm\_amenity \keyword{AS} T3 \keyword{ON} T2.amenid = T3.amenid \keyword{WHERE} T3.amenity\_name = `TV Lounge'}}
\\
\small{Q3: How many students are living there?}\\
\small{S3:} \scriptsize{\sql{\keyword{SELECT} \keyword{COUNT}(*) \keyword{FROM} student \keyword{AS} T1 \keyword{JOIN} lives\_in \keyword{AS} T2 \keyword{ON} T1.stuid = T2.stuid \keyword{WHERE} T2.dormid}}\\
\vspace{-.18in}
~~~~~\scriptsize{\sql{\keyword{IN} (\keyword{SELECT} T3.dormid \keyword{FROM} has\_amenity \keyword{AS} T3 \keyword{JOIN} dorm\_amenity \keyword{AS} T4 \keyword{ON} T3.amenid = T4.amenid}}\\
\vspace{-.23in}
~~~~~\scriptsize{\sql{\keyword{WHERE} T4.amenity\_name = `TV Lounge')}}
\\
\small{Q4: Please show their first and last names.}\\
\small{S4:} \scriptsize{\sql{\keyword{SELECT} T1.fname, T1.lname \keyword{FROM} student \keyword{AS} T1 \keyword{JOIN} lives\_in \keyword{AS} T2 \keyword{ON} T1.stuid = T2.stuid \keyword{WHERE} T2.dormid}}\\
\vspace{-.18in}
~~~~~\scriptsize{\sql{\keyword{IN} (\keyword{SELECT} T3.dormid \keyword{FROM} has\_amenity \keyword{AS} T3 \keyword{JOIN} dorm\_amenity \keyword{AS} T4 \keyword{ON} T3.amenid = T4.amenid}}\\
\vspace{-.24in}
~~~~~\scriptsize{\sql{\keyword{WHERE} T4.amenity\_name = `TV Lounge')}}\\
\hline
\end{tabular}
\caption{SParC example.}
\vspace{-3mm}
\label{tab:sparc_example}
\end{table*}

To exploit the correlation between sequentially generated queries and generalize the system to different domains, in this paper, we study an editing-based approach for cross-domain context-dependent text-to-SQL generation task.
We propose 
query generation by editing the query in the previous turn.
To this end, we first encode the previous query as a sequence of tokens, and the decoder computes a switch to change it at the token level.
This sequence editing mechanism 
models token-level changes and is thus robust to error propagation.
Furthermore, to capture the user utterance and the complex database schemas in different domains, we use an utterance-table encoder based on BERT to jointly encode the user utterance and column headers with co-attention, and adopt a table-aware decoder to perform SQL generation with attentions over both the user utterance and column headers.

We evaluate our model on SParC \cite{yu2019sparc}, a new large-scale dataset for cross-domain semantic parsing in context consisting of coherent question sequences annotated with SQL queries over 200 databases in 138 domains.
Experiment results show that by generating from the previous query, our model delivers an improvement of 7\% question match accuracy and 11\% interaction match accuracy over the previous state-of-the-art.
Further analysis shows that our editing approach is more robust to error propagation than copying segments, and the improvement becomes more significant if the basic text-to-SQL generation accuracy (without editing) improves.

\begin{figure*}[t!]
  \centering
  \includegraphics[width=\textwidth]{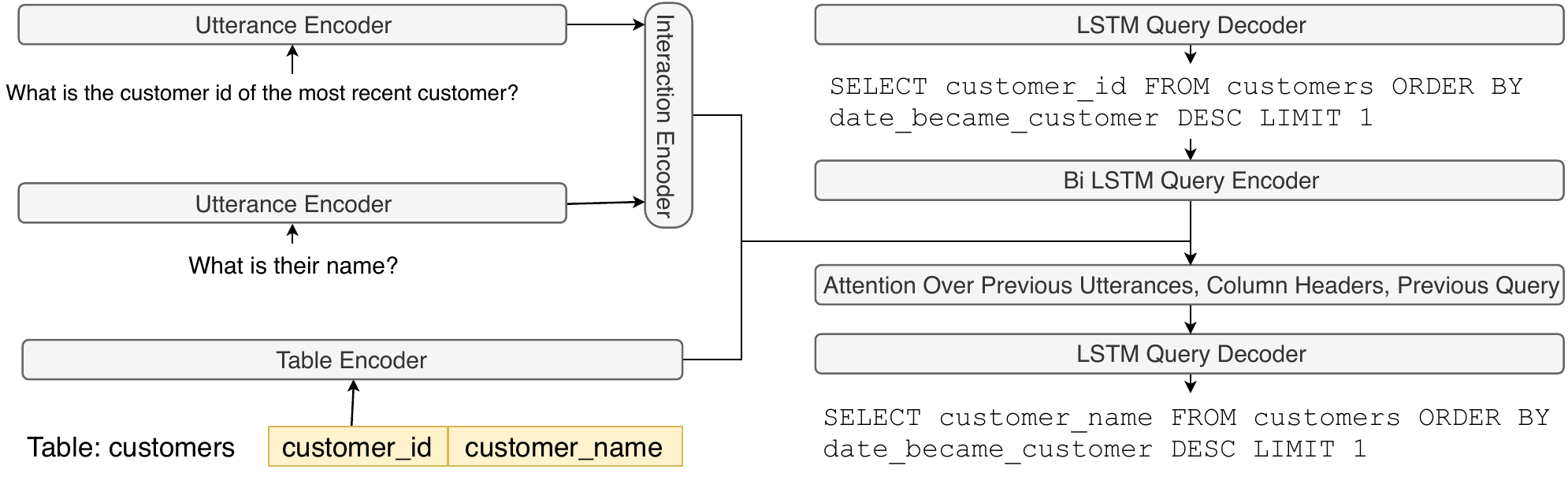}
  \caption{Model architecture of editing the previous query with attentions to the user utterances, the table schema, and the previously generated query.}
  \label{fig:edit_model}
\end{figure*}

\section{Cross-Domain Context-Depencent Semantic Parsing}
\subsection{Datasets}
We use \sparc\footnote{\url{https://yale-lily.github.io/sparc}} \cite{yu2019sparc}, a large-scale cross-domain context-dependent semantic parsing dataset with SQL labels, as our main evaluation benchmark.
A \sparc example is shown in Table \ref{tab:sparc_example}.
We also report performance on ATIS~\cite{hemphill1990atis,Dahl94} for direct comparison to~\newcite{suhr2018learning}.
In addition, we evaluate the cross-domain context-independent text-to-SQL ability of our model on Spider\footnote{\url{https://yale-lily.github.io/spider}} \cite{yu2018spider}, which SParC is built on.

We summarize and compare the data statistics in Table \ref{tab:dataset} and Table \ref{tab:dataset_2}.
While the ATIS dataset 
has been extensively studied, it is limited to a particular domain.
By contrast, \sparc is both context-dependent and cross-domain.
Each interaction in \sparc is constructed using a question in Spider as the interaction goal, where the annotator asks inter-related questions to obtain information that completes the goal.
\sparc contains interactions over 200 databases and it follows the same database split as Spider where each database appears only in one of train, dev and test sets.
In summary, \sparc introduces new challenges to context-dependent text-to-SQL because it (1) contains more complex context dependencies, (2) has greater semantic coverage, and (3) adopts a cross-domain task setting.

\subsection{Task Formulation}
Let $X$ denote a natural language utterance and $Y$ denote the corresponding SQL query.
Context-independent semantic parsing considers individual $(X,Y)$ pairs and maps $X$ to $Y$.
In context-dependent semantic parsing, we consider an interaction $I$ consisting of $n$ utterance-query pairs in a sequence: 
$$I = [(X_i,Y_i)]_{i=1}^{n}$$
At each turn $t$, the goal is to generate $Y_t$ given the current utterance $X_t$ and the interaction history 
$$[(X_1,Y_1),(X_2,Y_2),\dots,(X_{t-1},Y_{t-1})]$$

Furthermore, in the cross-domain setting, each interaction is grounded to a different database.
Therefore, the model is also given the schema of the current database as an input.
We consider relational databases with multiple tables, and each table contains multiple column headers:
$$T = [c_1,c_2,\dots,c_l,\dots,c_m]$$
where $m$ is the number of column headers, and each $c_l$ consists of multiple words including its table name and column name (\S~\ref{sec:utterance-table-encoder}).

\section{Methodology}
We employ an encoder-decoder architecture with attention mechanisms \cite{sutskever2014sequence,luong2015effective} as illustrated in Figure \ref{fig:edit_model}.
\hide{
While based on \newcite{suhr2018learning}, our model has differences including
}
The framework consists of (1) an utterance-table encoder 
to explicitly encode the user utterance and table schema at each turn, (2) A turn attention incorporating the recent history for decoding, (3) a table-aware decoder taking into account the context of the utterance, the table schema, and the previously generated query to make editing decisions.

\begin{figure*}[t!]
     \centering
     \begin{subfigure}[b]{0.45\textwidth}
         \centering
         \includegraphics[width=\textwidth]{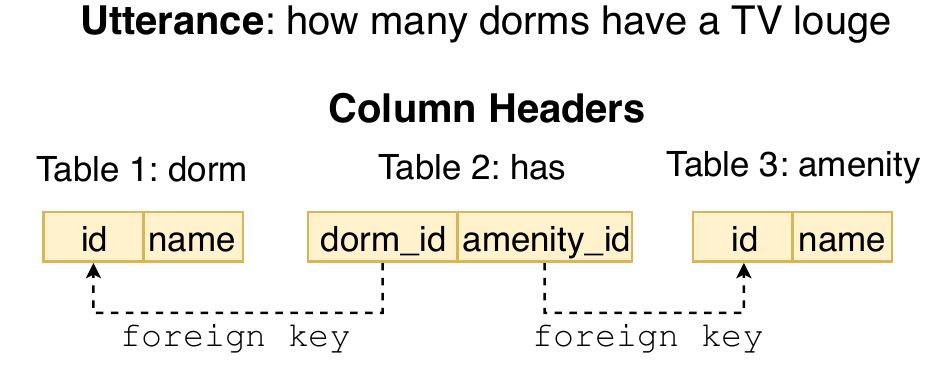}
         \caption{An example of user utterance and column headers.}
         \label{fig:example}
     \end{subfigure} %
     \hfill
     \begin{subfigure}[b]{0.45\textwidth}
         \centering
         \includegraphics[width=\textwidth]{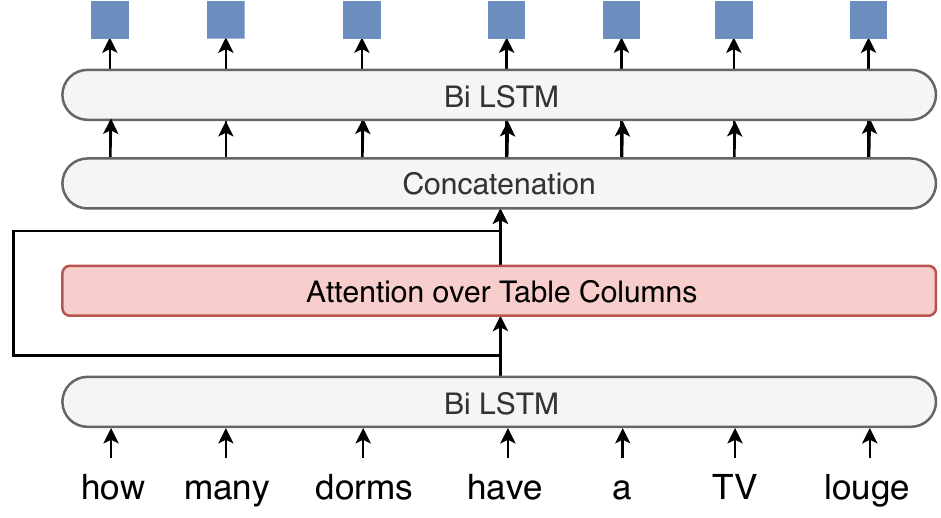}
         \caption{Utterance Encoder.}
         \label{fig:utterance_encoder}
     \end{subfigure} %
     \hfill
     \vfill
     \vspace{3mm}
     \begin{subfigure}[b]{\textwidth}
         \centering
         \includegraphics[width=\textwidth]{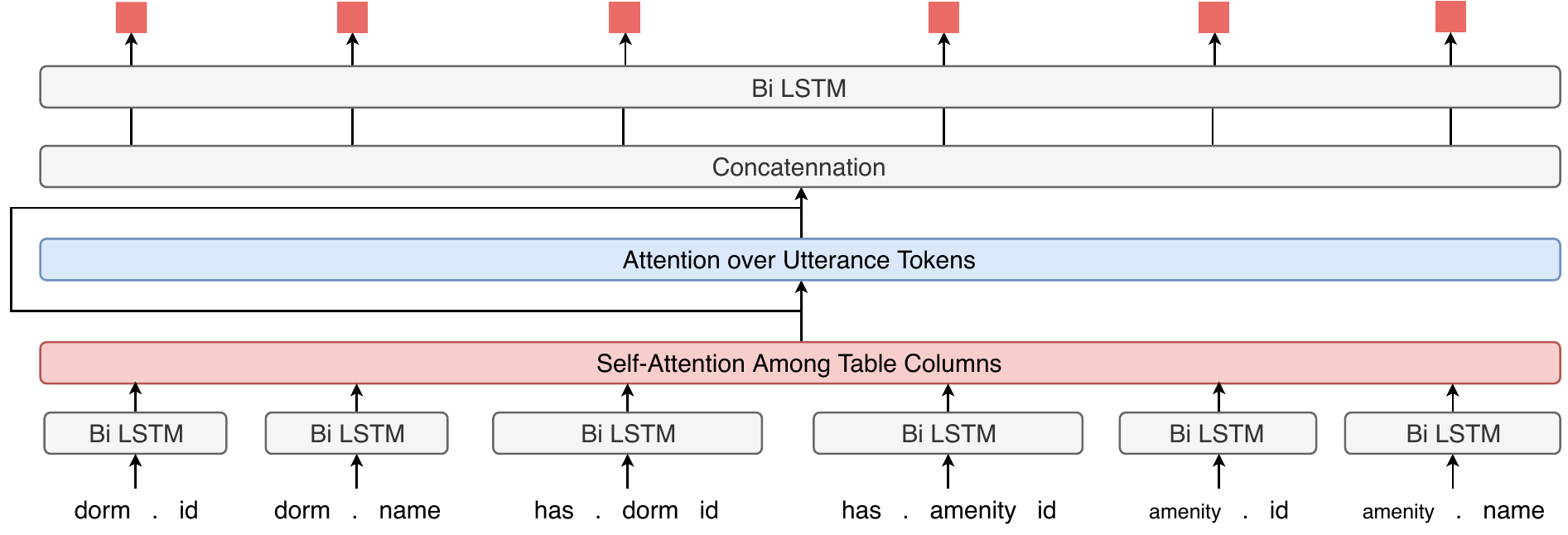}
         \caption{Table Encoder.}
         \label{fig:table_encoder}
     \end{subfigure}
\caption{Utterance-Table Encoder for the example in (a).}
\vspace{-3mm}
\label{fig:table_utterance_encoder}
\end{figure*}

\subsection{Utterance-Table Encoder}
\label{sec:utterance-table-encoder}
An effective encoder captures the meaning of user utterances, the structure of table schema, and the relationship between the two.
To this end, we build an utterance-table encoder with co-attention between the two as illustrated in Figure \ref{fig:table_utterance_encoder}.

Figure \ref{fig:utterance_encoder} shows the utterance encoder.
For the user utterance at each turn, we first use a bi-LSTM to encode utterance tokens.
The bi-LSTM hidden state is fed into a dot-product attention layer \cite{luong2015effective} over the column header embeddings.
For each utterance token embedding, we get an attention weighted average of the column header embeddings to obtain the most relevant columns \cite{dong2018coarse}.
We then concatenate the bi-LSTM hidden state and the column attention vector, and use a second layer bi-LSTM to generate the utterance token embedding $\newvec{h}^{E}$.

Figure \ref{fig:table_encoder} shows the table encoder.
For each column header, we concatenate its table name and its column name separated by a special dot token (i.e., \texttt{table name . column name}).
Each column header is processed by a bi-LSTM layer.
To better capture the internal structure of the table schemas (e.g., foreign key), we then employ a self-attention \cite{vaswani2017attention} among all column headers.
We then use an attention layer to capture the relationship between the utterance and the table schema.
We concatenate the self-attention vector and the utterance attention vector, and use a second layer bi-LSTM to generate the column header embedding $\newvec{h}^{C}$.

Note that the two embeddings depend on each other due to 
the co-attention, and thus the column header representation changes across different utterances in a single interaction.
\\\noindent
\textbf{Utterance-Table BERT Embedding}. We consider two options as the input to the first layer bi-LSTM.
The first choice is the pretrained word embedding.
Second, we also consider the contextualized word embedding based on BERT \cite{devlin2019bert}.
To be specific, we follow \newcite{hwang2019comprehensive} to concatenate the user utterance and all the column headers in a single sequence separated by the \texttt{[SEP]} token:
\begin{equation*}
    \texttt{[CLS]},X_i,\texttt{[SEP]}, c_1, \texttt{[SEP]}, \dots, c_m, \texttt{[SEP]}
\end{equation*}
This sequence is fed into the pretrained BERT model whose hidden states at the last layer is used as the input embedding.

\subsection{Interaction Encoder with Turn Attention}
\label{sec:turn-attention}
To capture the information across different utterances, we use an interaction-level encoder~\cite{suhr2018learning} on top of
the utterance-level encoder.
At each turn, we use the hidden state at the last time step from the utterance-level encoder as the utterance encoding.
This is the input to a uni-directional LSTM interaction encoder:
\begin{align*}
\begin{split}
    \newvec{h}^{U}_{i}   & = \newvec{h}^{E}_{i,|X_i|} \\
    \newvec{h}^{I}_{i+1} & = \text{LSTM}^{I}(\newvec{h}^{U}_{i}, \newvec{h}^{I}_{i}) \\
\end{split}
\end{align*}

The hidden state of this interaction encoder $\newvec{h}^{I}$ encodes the history as the interaction proceeds.
\\\noindent
\textbf{Turn Attention}
When issuing the current utterance, the user may omit or explicitly refer to the previously mentioned information.
\hide{
This requires the model to capture the relationships between the current utterance and the context history.
}
To this end, we adopt the turn attention mechanism to capture correlation between the current utterance and the utterance(s) at specific turn(s).
At the current turn $t$, we compute the turn attention by the dot-product attention between the current utterance and previous utterances in the history, and then add the weighted average of previous utterance embeddings to the current utterance embedding:
\begin{align}
\label{eq:turn_att}
\begin{split}
s_{i}                         & = \newvec{h}^{U}_{t}\newvec{W}_{\text{turn-att}}\newvec{h}^{U}_{i} \\
\alpha^{\text{turn}}          & = \softmax(s)  \\
\newvec{c}_{t}^{\text{turn}}  & = \newvec{h}^{U}_{t} + \sum_{i=1}^{t-1} \alpha_{i}^{\text{turn}} \times \newvec{h}^{U}_{i}
\end{split}
\end{align}
The $\newvec{c}_{t}^{\text{turn}}$ summarizes the context information and the current user query and will be used as the initial decoder state as described in the following.

\subsection{Table-aware Decoder}
\label{sec:table-aware-decoder}
We use an LSTM decoder with attention to generate SQL queries by incorporating the interaction history, the current user utterance, and the table schema.

Denote the decoding step as $k$, we provide the decoder input as a concatenation of the embedding of SQL query token $\newvec{q}_k$ and a context vector $\newvec{c}_k$:
\begin{equation*}
    \newvec{h}^{D}_{k+1} = \text{LSTM}^{D}([\newvec{q}_k;\newvec{c}_k], \newvec{h}^{D}_{k})
\end{equation*}
where $\newvec{h}^{D}$ is the hidden state of the decoder $\text{LSTM}^{D}$, and the hidden state $\newvec{h}^{D}_{0}$ is initialized by $\newvec{c}_{t}^{\text{turn}}$.
When the query token is a SQL keyword, $\newvec{q}_k$ is a learned embedding; when it is a column header, we use the column header embedding given by the table-utterance encoder as $\newvec{q}_k$.
The context vector $\newvec{c}_k$ is described below.
\\\noindent
\textbf{Context Vector with the Table and User Utterance}. 
The context vector consists of attentions to both the table and the user utterance.
First, at each step $k$, the decoder computes the attention between the decoder hidden state and the column header embedding.
\begin{align}
\label{eq:att1}
\begin{split}
s_{l}                         & = \newvec{h}^{D}_{k}\newvec{W}_{\text{column-att}}\newvec{h}^{C}_{l} \\
\alpha^{\text{column}}        & = \softmax(s)  \\
\newvec{c}_k^{\text{column}}  & = \sum_{l} \alpha_{l}^{\text{column}} \times \newvec{h}^{C}_{l} \\
\end{split}
\end{align}
where $l$ is the index of column headers and $\newvec{h}^{C}_{l}$ is its embedding.
Second, it also computes the attention between the decoder hidden state and the utterance token embeddings:
\begin{align}
\label{eq:att2}
\begin{split}
s_{i,j}                        & = \newvec{h}^{D}_{k}\newvec{W}_{\text{utterance-att}}\newvec{h}^{E}_{i,j} \\
\alpha^{\text{utterance}}      & = \softmax(s)  \\
\newvec{c}_k^{\text{token}}    & = \sum_{i,j} \alpha_{i,j}^{\text{utterance}} \times \newvec{h}^{E}_{i,j} \\
\end{split}
\end{align}
where $i$ is the turn index, $j$ is the token index, and $\newvec{h}^{E}_{i,j}$ is the token embedding for the $j$-th token of $i$-th utterance.
The context vector $\newvec{c}_k$ is a concatenation of the two:
\begin{equation*}
  \newvec{c}_k = [\newvec{c}_k^{\text{column}};\newvec{c}_k^{\text{token}}]  
\end{equation*}
\textbf{Output Distribution}. In the output layer, our decoder chooses to generate a SQL keyword (e.g., \texttt{
SELECT}, \texttt{WHERE}, \texttt{GROUP BY}, \texttt{ORDER BY}) or a column header.
This is critical for the cross-domain setting where the table schema changes across different examples.
To achieve this, we use separate layers to score SQL keywords and column headers, and finally use the softmax operation to generate the output probability distribution:
\begin{align}
\label{eq:output}
\begin{split}
\newvec{o}_{k}             & = \tanh([\newvec{h}^{D}_{k};\newvec{c}_k]\newvec{W}_{o}) \\
\newvec{m}^{\text{SQL}}    & = \newvec{o}_{k}\newvec{W}_{\text{SQL}} + \newvec{b}_{\text{SQL}} \\
\newvec{m}^{\text{column}} & = \newvec{o}_{k}\newvec{W}_{\text{column}}\newvec{h}^{C} \\
P(y_k)                     & = \softmax([\newvec{m}^{\text{SQL}}; \newvec{m}^{\text{column}}]) \\
\end{split}
\end{align}

\begin{figure}[t!]
  \centering
  \includegraphics[width=.93\columnwidth]{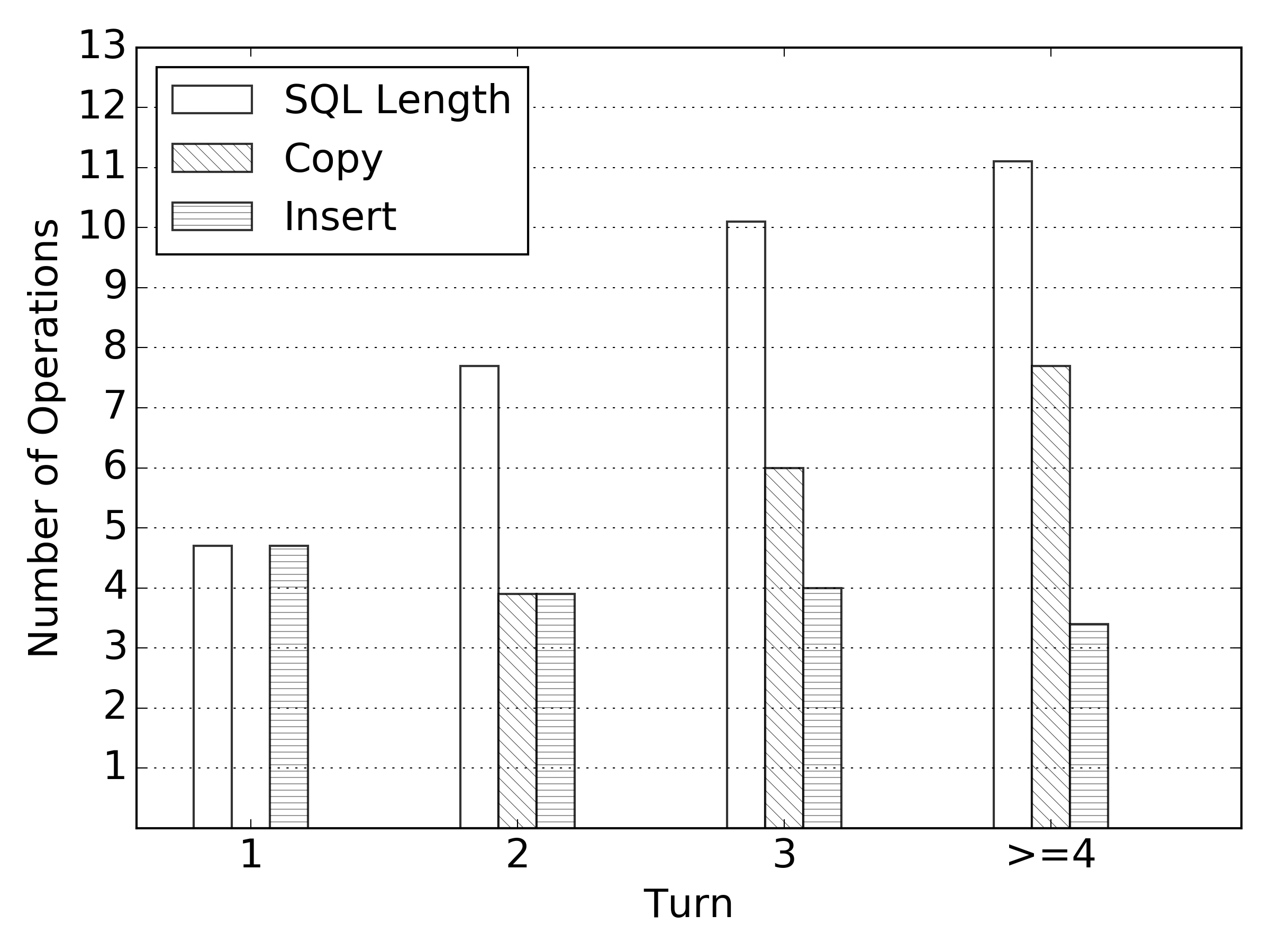}
  \caption{Number of operations at different turns.}
  \vspace{-4mm}
  \label{fig:num_operation}
\end{figure}

\vspace{-2mm}
\subsection{Query Editing Mechanism}
\label{sec:query_edit}
In an interaction with the system, the user often asks a sequence of closely related questions to complete the final query goal.
Therefore, the query generated for the current turn often overlaps significantly with the previous ones.

To empirically verify the usefulness of leveraging the previous query, 
we consider the process of generating the current query by applying copy and insert operations to the previous query\footnote{We use a diffing algorithm from \url{https://github.com/paulgb/simplediff}}.
Figure \ref{fig:num_operation} shows the SQL query length and the number of copy and insert operations at different turns.
As the interaction proceeds, the user question becomes more complicated as it requires longer SQL query to answer.
However, more query tokens overlap with the previous query, and thus the number of new tokens remains small at the third turn and beyond.

Based on this observation, we extend our table-ware decoder with a query editing mechanism.
We first encode the previous query using another bi-LSTM, and its hidden states are the query token embeddings $\newvec{h}^{Q}_{i,j^{\prime}}$ (i.e., the $j^{\prime}$-th token of the $i$-th query).
We then extend the context vector with the attention to the previous query:
\begin{equation*}
  \newvec{c}_k = [\newvec{c}_k^{\text{column}};\newvec{c}_k^{\text{token}};\newvec{c}_k^{\text{query}}]  
\end{equation*}
where $\newvec{c}_k^{\text{query}}$ is produced by an attention to query tokens $\newvec{h}^{Q}_{i,j^{\prime}}$ in the same form as Equation \ref{eq:att2}.

At each decoding step, we predict a switch $p_{\text{copy}}$ to decide if we need copy from the previous query or insert a new token.
\begin{align}
\label{eq:edit}
\begin{split}
p_{\text{copy}}    & = \sigma(\newvec{c}_{k}\newvec{W}_{\text{copy}} + \newvec{b}_{\text{copy}}) \\
p_{\text{insert}}  & = 1 - p_{\text{copy}} \\
\end{split}
\end{align}
Then, we use a separate layer to score the query tokens at turn $t-1$, and the output distribution is modified as the following to take into account the editing probability:

\begin{align}
\label{eq:edit2}
\begin{split}
P_\text{prev\_SQL} & = \softmax(\newvec{o}_{k}\newvec{W}_{\text{prev\_SQL}}\newvec{h}^{Q}_{t-1}) \\
\newvec{m}^{\text{SQL}} & = \newvec{o}_{k}\newvec{W}_{\text{SQL}} + \newvec{b}_{\text{SQL}} \\
\newvec{m}^{\text{column}} & = \newvec{o}_{k}\newvec{W}_{\text{column}}\newvec{h}^{C} \\
P_{\text{SQL} \bigcup \text{column}} & = \softmax([\newvec{m}^{\text{SQL}}; \newvec{m}^{\text{column}}]) \\
P(y_k) & = p_{\text{copy}} \cdot P_\text{prev\_SQL}(y_k \in \text{prev\_SQL}) \\
 + p_{\text{insert}} & \cdot P_{\text{SQL} \bigcup \text{column}}(y_k \in \text{SQL} \bigcup \text{column})
\end{split}
\end{align}

While the copy mechanism has been introduced by \newcite{gu2016incorporating} and \newcite{see2017get}, they focus on summarization or response generation applications by copying from the source sentences.
By contrast, our focus is on editing the previously generated query while incorporating the context of user utterances and table schemas. 

\vspace{-2mm}
\section{Related Work}
\vspace{-2mm}
Semantic parsing is the task of mapping natural language sentences into formal representations.
It has been studied for decades including using linguistically-motivated compositional representations, such as logical forms \cite{zelle96,clarke2010driving} and lambda calculus \cite{zettlemoyer2005learning,artzi2011bootstrapping}, and using executable programs, such as SQL queries \cite{miller1996fully,zhong2017seq} and other general-purpose programming languages \cite{yin2017syntactic,iyer2018mapping}.
Most of the early studies worked on a few domains and small datasets such as GeoQuery \cite{zelle96} and Overnight \cite{wang2015building}.

Recently, large and cross-domain text-to-SQL datasets such as WikiSQL \cite{zhong2017seq} and Spider \cite{yu2018spider} have received an increasing amount of attention as many data-driven neural approaches achieve promising results \cite{dong2016language, su2017cross, iyer2017learning, xu2017sqlnet, dollak2018improving, yu2018typesql, huang2018natural, dong2018coarse, sun2018semantic, gur2018dialsql, guo2018question, yavuz2018takes, shi2018incsql}.
Most of them still focus on \textit{context-independent} semantic parsing by converting single-turn questions into executable queries.

Relatively less effort has been devoted to \textit{context-dependent} semantic parsing on datasets including ATIS \cite{hemphill1990atis,dahl1994expanding}, SpaceBook \cite{vlachos2014new}, SCONE \cite{long2016simpler,guu2017language,fried2018unified,suhr2018situated,huang2019flowqa}, SequentialQA \cite{iyyer2017search}, SParC \cite{yu2019sparc} and CoSQL \cite{yu2019cosql}.
On ATIS, \newcite{miller1996fully} maps utterances to semantic frames which are then mapped to SQL queries; \newcite{zettlemoyer2009learning} starts with context-independent Combinatory Categorial Grammar (CCG) parsing and then resolves references to generate lambda-calculus logical forms for sequences of sentences.
The most relevant to our work is \newcite{suhr2018learning}, who generate ATIS SQL queries from interactions by incorporating history with an interaction-level encoder and copying segments of previously generated queries.
Furthermore, SCONE contains three domains using stack- or list-like elements and most queries include a single binary predicate.
SequentialQA is created by decomposing some complicated questions in WikiTableQuestions \cite{pasupat2015compositional}.
Since both SCONE and SequentialQA are annotated with only denotations but not query labels, they don't include many questions with rich semantic and contextual types.
For example, SequentialQA \cite{iyyer2017search} requires that the answer to follow-up questions must be a subset of previous answers, and most of the questions can be answered by simple SQL queries with \texttt{SELECT} and \texttt{WHERE} clauses.

Concurrent with our work, \newcite{yu2019cosql} introduced CoSQL, a large-scale cross-domain conversational text-to-SQL corpus collected under the  Wizard-of-Oz setting. Each dialogue in CoSQL simulates a DB querying scenario with a crowd worker as a user and a college computer science student who is familiar with SQL as an expert. Question-SQL pairs in CoSQL reflect greater diversity in user backgrounds compared to other corpora and involve frequent changes in user intent between pairs or ambiguous questions that require user clarification. These features pose new challenges for text-to-SQL systems.

Our work is also related to recently proposed approaches to code generation by editing \cite{hayati2018retrieval,yin2019learning,hashimoto2018retrieve}.
While they follow the framework of generating code by editing the relevant examples retrieved from training data, we focus on a context-dependent setting where we generate queries from the previous query predicted by the system itself.

\section{Experimental Results}
\vspace{-1mm}
\subsection{Metrics}
\vspace{-1mm}
On both Spider and SParC, we use the exact set match accuracy between the gold and the predicted queries \footnote{More details at \scriptsize \url{https://github.com/taoyds/spider/tree/master/evaluation_examples}}.
To avoid ordering issues, instead of using simple string matching, \newcite{yu2018spider} decompose predicted queries into different SQL clauses such as \texttt{SELECT}, \texttt{WHERE}, \texttt{GROUP BY}, and \texttt{ORDER BY} and compute scores for each clause using set matching separately.
On SparC, we report two metrics: question match accuracy which is the score average over all questions and interaction match accuracy which is average over all interactions.

\vspace{-1mm}
\subsection{Baselines}
\vspace{-1mm}

\textbf{SParC.}
We compare with the two baseline models released by \newcite{yu2019sparc}.

(1) Context-dependent Seq2Seq (CD-Seq2Seq): This model is adapted from \newcite{suhr2018learning}. The original model was developed for ATIS and does not take the database schema as input hence cannot generalize well across domains. \newcite{yu2019sparc} adapt it to perform context-dependent SQL generation in multiple domains by adding a bi-LSTM database schema encoder which takes bag-of-words representations of column headers as input. They also modify the decoder to select between a SQL keyword or a column header.

(2) SyntaxSQL-con: This is adapted from the original context-agnostic SyntaxSQLNet~\cite{yu2018syntaxsqlnet} by using bi-LSTMs to encode the interaction history including the utterance and the associated SQL query response.
It also employs a column attention mechanism to compute representations of the previous question and SQL query.

\textbf{Spider.}
We compare with the results as reported in \newcite{yu2018syntaxsqlnet}.
Furthermore, we also include recent results from \newcite{lee2019recursive} who propose to use recursive decoding procedure, \newcite{bogin2019representing} introducing graph neural networks for encoding schemas, and \newcite{guo2019towards} who achieve state-of-the-art performance by using an intermediate representation to bridge natural language questions and SQL queries.

\subsection{Implementation Details}
Our model is implemented in PyTorch \cite{paszke2017automatic}.
We use pretrained 300-dimensional GloVe \cite{pennington2014glove} word embedding.
All LSTM layers have 300 hidden size, and we use 1 layer for encoder LSTMs, and 2 layers for decoder LSTMs.
We use the ADAM optimizer \cite{kingma2015adam} to minimize the token-level cross-entropy loss with a batch size of 16.
Model parameters are randomly initialized from a uniform distribution $U[-0.1,0.1]$.
The main model has an initial learning rate of 0.001 and it will be multiplied by 0.8 if the validation loss increases compared with the previous epoch.
When using BERT instead of GloVe, we use the pretrained small uncased BERT model with 768 hidden size\footnote{\scriptsize \url{https://github.com/google-research/bert}}, and we fine tune it with a separate constant learning rate of 0.00001.
The training typically converges in 10 epochs.

\begin{table}[t!]
\resizebox{\columnwidth}{!}{
\begin{tabular}{lcc}
\Xhline{4\arrayrulewidth}
                                         & Dev Set & Test Set \\ \hline
SQLNet \cite{xu2017sqlnet}                & 10.9 & 12.4\\
SyntaxSQLNet \cite{yu2018syntaxsqlnet}    & 18.9 & 19.7\\
\quad+data augmentation \cite{yu2018syntaxsqlnet}   & 24.8 & 27.2\\
\newcite{lee2019recursive}                & 28.5 & 24.3 \\
GNN \cite{bogin2019representing}           & 40.7 & 39.4\\
IRNet \cite{guo2019towards}                  & 53.2 & 46.7\\
IRNet (BERT) \cite{guo2019towards}            & 61.9 & 54.7\\ \hline
Ours                                       & 36.4 & 32.9\\
\quad + utterance-table BERT Embedding     & 57.6 & 53.4\\
\Xhline{4\arrayrulewidth}
\end{tabular}
}
\caption{Spider results on dev set and test set.
}
\vspace{-3mm}
\label{tab:spider_result}
\end{table}

\begin{table*}[th!]
\centering
\resizebox{0.95\textwidth}{!}{
\begin{tabular}{lcccc}
\Xhline{4\arrayrulewidth}
    & \multicolumn{2}{c}{Question Match} & \multicolumn{2}{c}{Interaction Match} \\
              & Dev & Test &  Dev & Test \\ \hline
\syncon{} \cite{yu2019sparc}            & 18.5 & 20.2 & 4.3 & 5.2 \\
\seqcon{} \cite{yu2019sparc}$^*$       & 21.9 & 23.2 & 8.1 & 7.5 \\
 \quad + segment copy (w/ predicted query)   & 21.7 & 20.3 & 9.5 & 8.1 \\
 \quad + segment copy (w/ gold query)        & 27.3 & 26.7 & 10.0 & 8.4  \\
\hline
Ours                                      & 31.4 & -- & 14.7 & -- \\
 \quad + query attention and sequence editing (w/ predicted query) & 33.0 & -- & 16.4 & -- \\
 \quad + query attention and sequence editing (w/ gold query)      & 40.6 & -- & 17.3 & -- \\
Ours + utterance-table BERT Embedding         & 40.4 & -- & 18.1 & -- \\
 \quad + query attention (w/ predicted query)                      & 42.7 &  --  & 21.6 & -- \\
 \quad + query attention and sequence editing (w/ predicted query) & \textbf{47.2} & \textbf{47.9} & \textbf{29.5} & \textbf{25.3} \\
 \quad + query attention and sequence editing (w/ gold query)      & 53.4 & 54.5 & 29.2 & 25.0  \\
 
\Xhline{4\arrayrulewidth}
\end{tabular}
}
\caption{SParC results. For our models, we only report test set results of our best model on the dev set.
$^*$We improve the \seqcon{} performance over \newcite{yu2019sparc} by separating and parsing the column names (e.g., stu\_fname $\rightarrow$ student first name) and using the schema-specific output vocabulary during decoding.
}
\vspace{-2mm}
\label{tab:sparc_result}
\end{table*}

\begin{table*}[ht!]
\centering
\resizebox{0.85\textwidth}{!}{
\begin{tabular}{@{\extracolsep{8pt}}lcccccc@{}}
\Xhline{4\arrayrulewidth}
      & \multicolumn{3}{c}{Dev Set}   &    \multicolumn{3}{c}{Test Set} \\ \cline{2-4} \cline{5-7}
 & Query & Relaxed & Strict & Query & Relaxed & Strict \\ \hline
FULL \cite{suhr2018learning}    & 37.5$_{\pm0.9}$        & 63.0$_{\pm0.7}$ & 62.5$_{\pm0.9}$   & 43.6$_{\pm1.0}$        & 69.3$_{\pm0.8}$ & 69.2$_{\pm0.8}$            \\
Ours  & 36.2 & 60.5 & 60.0 & 43.9 & 68.5 & 68.1 \\
\Xhline{4\arrayrulewidth}
\end{tabular}
}
\caption{ATIS results on dev set and test set.}
\vspace{-3mm}
\label{tab:atis_result}
\end{table*}

\subsection{Overall Results}
\textbf{Spider.}
Table \ref{tab:spider_result} shows the results on Spider dataset.
Since each question is standalone, we don't use interaction-level decoder or query editing.
Our method can achieve the performance of 36.4\% on dev set and 32.9\% on test set, serving as a strong model for the context-independent cross-domain text-to-SQL generation.
This demonstrates the effectiveness of our utterance-table encoder and table-aware decoder to handle the semantics of user utterances and the complexity of table schemas to generate complex SQL queries in unseen domains.

Furthermore, adding the utterance-table BERT embedding gives significant improvement, achieving 57.6\% on dev set and 53.4\% on test set, which is comparable to the state-of-the-art results from IRNet with BERT.
We attribute our BERT model's high performance to (1) the empirically powerful text understanding ability of pretrained BERT model and (2) the early interaction between utterances and column headers when they are concatenated in a single sequence as the BERT input.

\begin{figure*}[t!]
     \centering
     \begin{subfigure}[t]{0.46\textwidth}
         \centering
         \includegraphics[width=\textwidth]{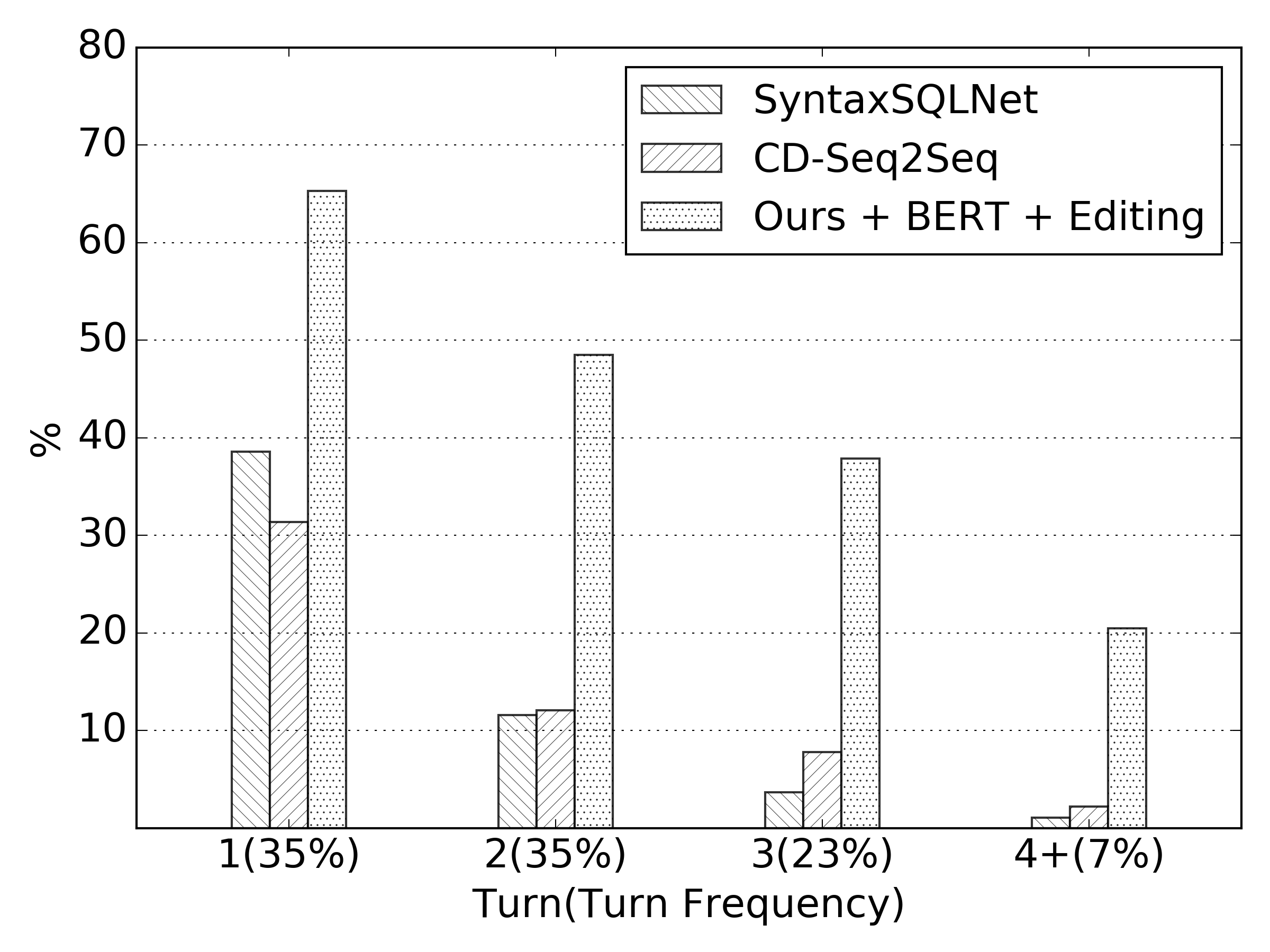}
         \vspace{-2mm}
         \label{fig:turn_acc}
     \end{subfigure}%
     \begin{subfigure}[t]{0.46\textwidth}
         \centering
         \includegraphics[width=\textwidth]{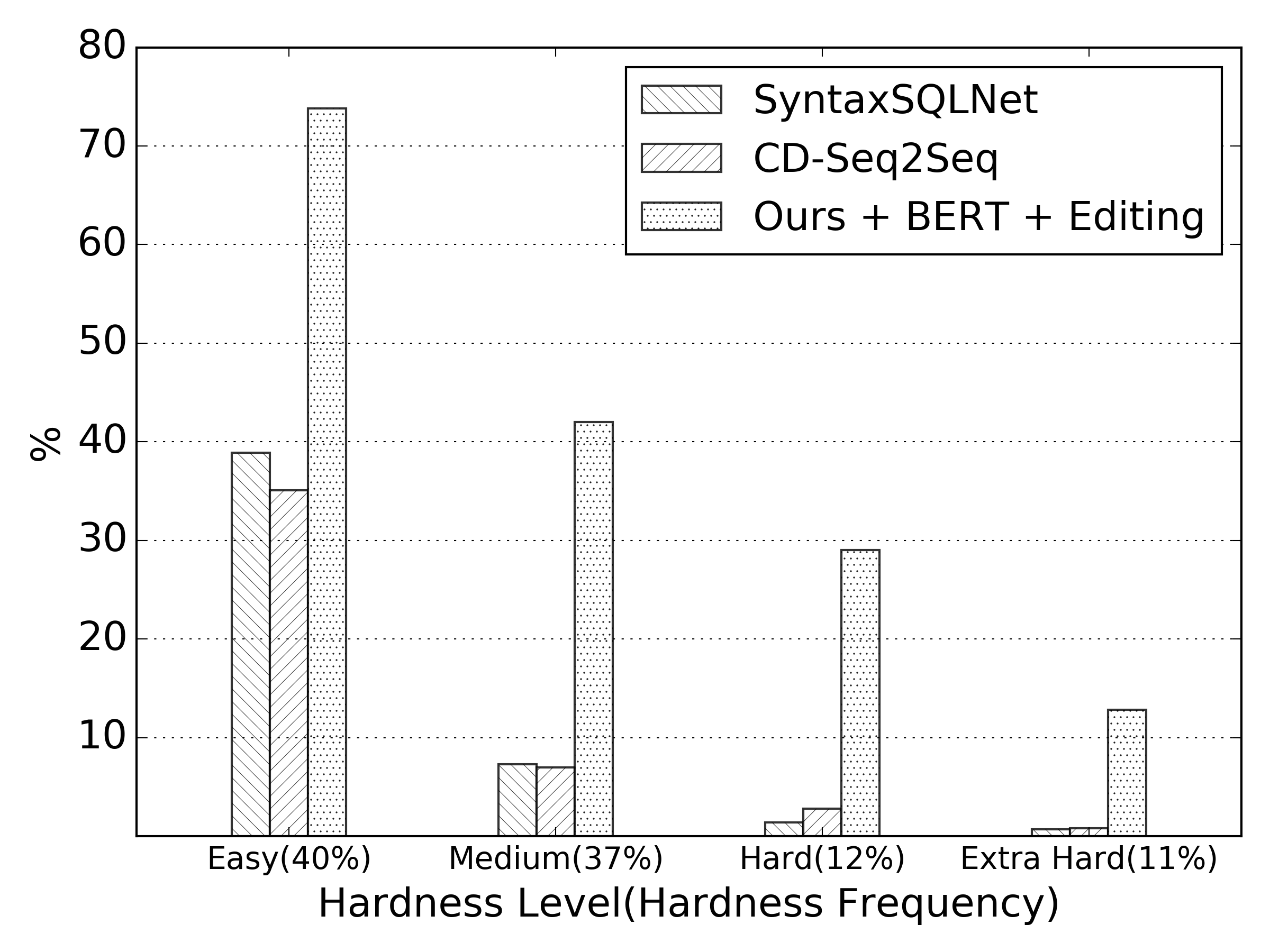}
         \vspace{-2mm}
         \label{fig:hardness_acc}
     \end{subfigure}
     \vspace{-5mm}
\caption{Performance split by different turns (Left) and hardness levels (Right) on SParC dev set.}
\vspace{-4mm}
\label{fig:sparc_dev}
\end{figure*}

\begin{figure}[t!]
  \centering
  \includegraphics[width=.90\columnwidth]{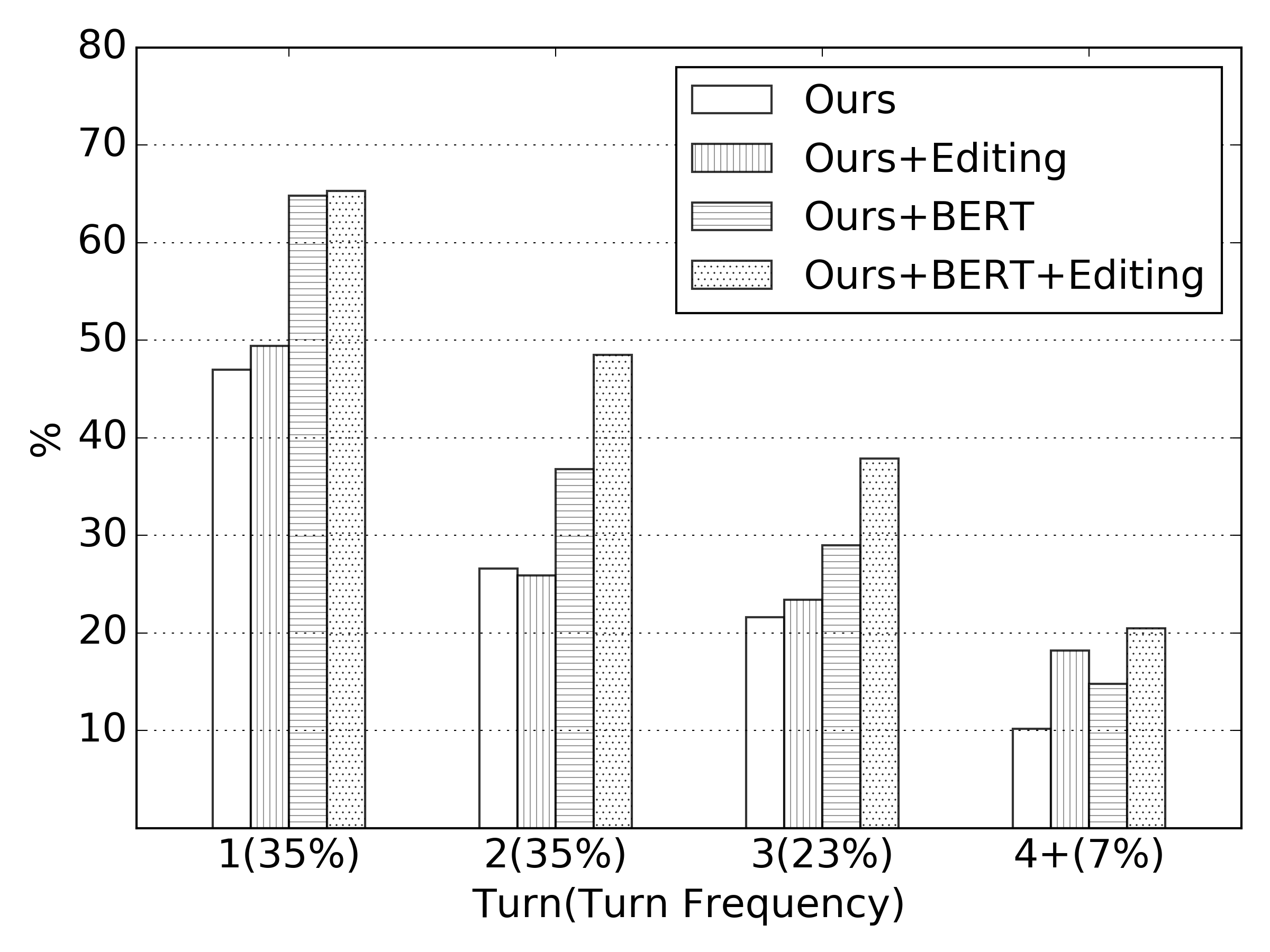}
  \caption{Effect of query editing at different turns on SParC dev set.}
  \vspace{-5mm}
  \label{fig:turn_acc_edit}
\end{figure}

\textbf{SParC.}
Table \ref{tab:sparc_result} shows the results on SParC dataset.
Similar to Spider, our model without previous query as input already outperforms \syncon{}, achieving 31.4\% question matching accuracy and 14.7\% interaction matching accuracy.
In addition, compared with CD-Seq2Seq, our model enjoys the benefits of the table-utterance encoder, turn attention, and the joint consideration of utterances and table schemas during the decoding stage.
This boosts the performance by 10\% question accuracy and 6\% interaction accuracy.

Furthermore, we also investigate the effect of copying segment.
We use the same segment copy procedure as \newcite{suhr2018learning}: first deterministically extract segments from the previous query and encode each segment using an LSTM, then generate a segment by computing its output probability based on its segment encoding.
However, since the segment extraction from \newcite{suhr2018learning} is exclusively designed for the ATIS dataset, we implement our own segment extraction procedure by extracting \texttt{SELECT}, \texttt{FROM}, \texttt{GROUP BY}, \texttt{ORDER BY} clauses as well as different conditions in \texttt{WHERE} clauses.
In this way, 3.9 segments can be extracted per SQL on average.
We found that adding segment copying to CD-Seq2Seq gives a slightly lower performance on question matching and a small gain on interaction matching, while using segments extracted from the gold query can have much higher results.
This demonstrates that segment copy is vulnerable to error propagation. In addition, it can only copy whole segments hence has difficulty capturing the changes of only one or a few tokens in the query.

To better understand how models perform as the interaction proceeds, Figure \ref{fig:sparc_dev} (Left) shows the performance split by turns on the dev set.
The questions asked in later turns are more difficult to answer given longer context history.
While the baselines have lower performance as the turn number increases, our model still maintains 38\%-48\% accuracy for turn 2 and 3, and 20\% at turn 4 or beyond.
Similarly, Figure \ref{fig:sparc_dev} (Right) shows the performance split by hardness levels with the frequency of examples.
This also demonstrates our model is more competitive in answering hard and extra hard questions.

\textbf{ATIS.}
We also report our model performance on ATIS in Table \ref{tab:atis_result}.
Our model achieves 36.2\% dev and 43.9\% test string accuracy, comparable to \newcite{suhr2018learning}.
On ATIS, we only apply our editing mechanism and reuse their utterance encoder instead of the BERT utterance-table encoder, because ATIS is single domain.

\subsection{Effect of Query Editing}
We further investigate the effect of our query editing mechanism.
To this end, we apply editing from both the gold query and the predicted query on our model with or without the utterance-table BERT embedding.
We also perform an ablation study to validate the contribution of query attention and sequence editing separately.

As shown in Table \ref{tab:sparc_result}, editing the gold query consistently improves both question match and interaction match accuracy.
This shows the editing approach is indeed helpful to improve the generation quality when the previous query is the oracle.

Using the predicted query is a more realistic setting, and in this case, the model is affected by error propagation due to the incorrect queries produced by itself.
For the model without the utterance-table BERT embedding, using the predicted query only gives around 1.5\% improvement.
As shown in Figure \ref{fig:turn_acc_edit}, this is because the editing mechanism is more helpful for turn 4 which is a small fraction of all question examples.
For the model with the utterance-table BERT embedding, the query generation accuracy at each turn is significantly improved, thus reducing the error propagation effect.
In this case, the editing approach delivers consistent improvements of 7\% increase on question matching accuracy and 11\% increase on interaction matching accuracy.
Figure \ref{fig:turn_acc_edit} also shows that query editing with BERT benefits all turns.

Finally, as an ablation study, Table \ref{tab:sparc_result} also reports the result with only query attention (use predicted query) on the dev set.
This improves over our vanilla BERT model without query attention and achieves 42.7\% question and 21.6\% interaction matching accuracy.
With query editing, our best model further improves to 47.2\% question and 29.5\% interaction matching accuracy.
This demonstrates the effectiveness of our query attention and query editing separately, both of which are essential to make use of the previous query.

\vspace{-2mm}
\section{Conclusions}
\vspace{-2mm}
In this paper, we propose an editing-based encoder-decoder model to address the problem of context-dependent cross-domain text-to-SQL generation.
While being simple, empirical results demonstrate the benefits of our editing mechanism.
The approach is more robust to error propagation than copying segments, and its performance increases when the basic text-to-SQL generation quality (without editing) is better.

\section*{Acknowledgements}
We thank the anonymous reviewers for their thoughtful detailed comments.

\bibliography{emnlp-ijcnlp-2019}
\bibliographystyle{acl_natbib}

\end{document}